\documentclass[10pt,twocolumn,letterpaper]{article}

\usepackage[usenames,dvipsnames]{xcolor} 
\usepackage{cvpr}
\usepackage{times}
\usepackage{amsmath}
\usepackage{amssymb}
\usepackage{mathtools}
\usepackage{bigints}
\usepackage{cite} 
\usepackage{authblk} 
\usepackage[caption=false,font=scriptsize,justification=raggedright,singlelinecheck=true]{subfig} 
\usepackage{url}
\usepackage[multidot]{grffile} 
\usepackage{graphicx}
\usepackage{glossaries}
\usepackage{algpseudocode} 

\def \R{{\mathbb{R}}}

\def \B{{\bar{B}}}

\def \alg1 {algorithm \ref{alg1}}

\newcommand{\comment}[1]{}

\newcommand{\matris}[1]{ \left[ \begin{smallmatrix} #1 \end{smallmatrix} \right]}

\newcommand{\argmin}{\operatornamewithlimits{arg\ min}}
\newcommand{\argmax}{\operatornamewithlimits{arg\ max}}

\def \L{{\mathcal{L}}}

\makeglossaries
\newacronym{ACFR}{ACFR}{Australian centre for field robotics}
\newacronym{ACRV}{ACRV}{Australian centre for robotic vision}

\newacronym{AUV}{AUV}{autonomous underwater vehicle}
\newacronym{UAV}{UAV}{unmanned aerial vehicle}
\newacronym{USV}{USV}{unmanned surface vehicle}
\newacronym{UGV}{UGV}{unmanned ground vehicle}
\newacronym{GPS}{GPS}{global positioning system}
\newacronym{SLAM}{SLAM}{simultaneous localisation and mapping}
\newacronym{IMU}{IMU}{inertial measurement unit}

\newacronym{MDSP}{MDSP}{multi-dimensional signal processing}
\newacronym{ROS}{ROS}{region of support}
\newacronym{DOF}{DOF}{degree-of-freedom}
\newacronym{RMS}{RMS}{root mean square}
\newacronym{SNR}{SNR}{signal-to-noise ratio}
\newacronym{CNR}{CNR}{contrast-to-noise ratio}
\newacronym{PCA}{PCA}{principal component analysis}

\newacronym{FIR}{FIR}{finite impulse response}
\newacronym{IIR}{IIR}{infinite impulse response}
\newacronym{DFT}{DFT}{discrete Fourier transform}
\newacronym{FFT}{FFT}{fast Fourier transform}
\newacronym{PSNR}{PSNR}{peak signal-to-noise ratio}
\newacronym{FPGA}{FPGA}{field programmable gate array}
\newacronym{GPU}{GPU}{graphics processing unit}
\newacronym{ASIC}{ASIC}{application-specific integrated circuit}
\newacronym{BW}{BW}{bandwidth}

\newacronym{PSF}{PSF}{point spread function}
\newacronym{FOV}{FOV}{field of view}
\newacronym{BRDF}{BRDF}{bidirectional reflectance distribution function}
\newacronym{FWHM}{FWHM}{full width at half maximum}

\newacronym{RANSAC}{RANSAC}{random sampling and consensus}

\newacronym{RL}{RL}{Richardson-Lucy}

\usepackage[ colorlinks=true, citecolor=Green, linkcolor=Red, urlcolor=black, bookmarks=true, bookmarksnumbered=true, pdftex, plainpages=false, pdfpagelabels]{hyperref}

\def \L{{\mathcal{L}}}

\cvprfinalcopy 

\def\cvprPaperID{0006} 

\ifcvprfinal\fi
\begin{document}

\title{Richardson-Lucy Deblurring for Moving Light Field Cameras}

\author{Donald G. Dansereau$^1$, Anders Eriksson$^2$ and J\"urgen Leitner$^{2,3}$\\
$^1$Stanford University, $^2$Queensland University of Technology, Brisbane, Australia\\
$^3$ARC Centre of Excellence for Robotic Vision, Brisbane, Australia\\
{\tt \small donald.dansereau@gmail.com, \{anders.eriksson, j.leitner\} @qut.edu.au}}


\maketitle

\begin{abstract}
We generalize Richardson-Lucy (RL) deblurring to 4-D light fields by replacing the convolution steps with light field rendering of motion blur. The method deals correctly with blur caused by 6-degree-of-freedom camera motion in complex 3-D scenes, without performing depth estimation.  We introduce a novel regularization term that maintains parallax information in the light field while reducing noise and ringing.  We demonstrate the method operating effectively on rendered scenes and scenes captured using an off-the-shelf light field camera. An industrial robot arm provides repeatable and known trajectories, allowing us to establish quantitative performance in complex 3-D scenes. Qualitative and quantitative results confirm the effectiveness of the method, including commonly occurring cases for which previously published methods fail. We include mathematical proof that the algorithm converges to the maximum-likelihood estimate of the unblurred scene under Poisson noise. We expect extension to blind methods to be possible following the generalization of 2-D Richardson-Lucy to blind deconvolution.
\end{abstract}

\section{Introduction}

The tradeoff between light gathering and sensitivity to motion blur makes effective image capture in low light or on mobile platforms difficult.  This is commonly an issue in robotics applications, e.g. \gls{UAV} and \gls{AUV} deployments in which cameras are in constant motion and light is often limited.   Handheld photography is also affected, especially on low-end cameras with low light sensitivity, but also on higher-end devices operating in low-light scenarios.

\begin{figure}
	\centering
	\includegraphics[width=1\columnwidth]{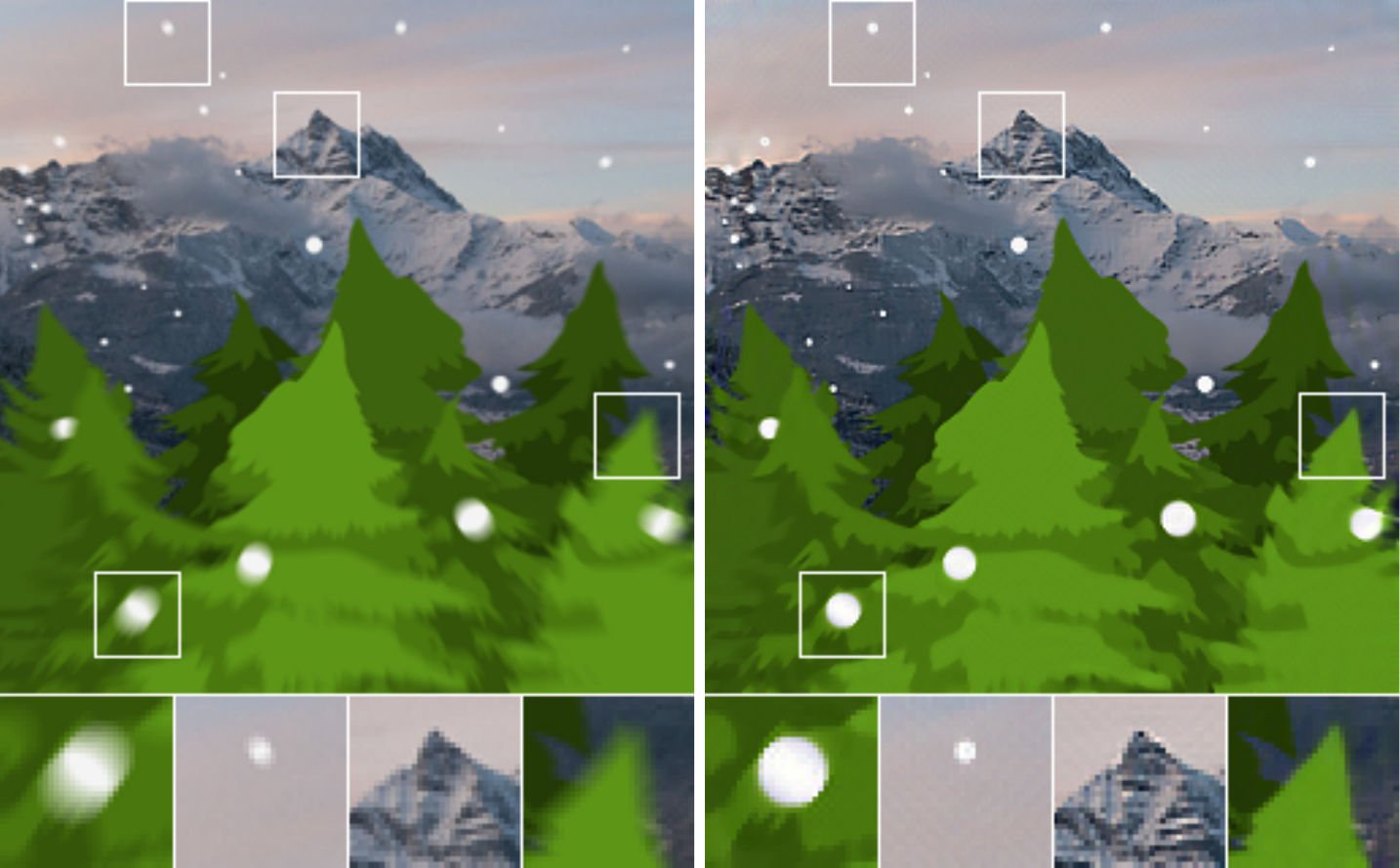}
	\caption{(left) Motion blur in 3-D scenes takes on a complex variety of shapes; (right) We introduce a light-field generalization of Richardson-Lucy deblurring which deals correctly with complex 3-D geometry and 6-DOF camera motion. No depth estimation is performed, only the camera's trajectory is required.}
	\label{Fig_VisualAbstract}
\end{figure}

The possibility of removing blur post-capture is enticing, and deblurring is a well-explored topic with previous work addressing the cases of spatially invariant blur~\cite{richardson1972bayesian,lucy1974iterative,chandramouli2014light} or planar projective motion~\cite{tai2011richardson}.  These approaches have in common that they do not apply to general 3-D scenes, where parallax motion results in a complex scene-dependent spatially varying blur kernel --~see Fig.~\ref{Fig_VisualAbstract} for example. Previous generalizations to light fields have similarly restricted scene geometry~\cite{chandramouli2014light}, or restricted camera motion to a plane and relied on explicit 3-D shape estimation, a potentially error-prone process in the case of a blurry input~\cite{snoswell2014light}.

\begin{figure}
	\centering
	\includegraphics[width=0.5\columnwidth]{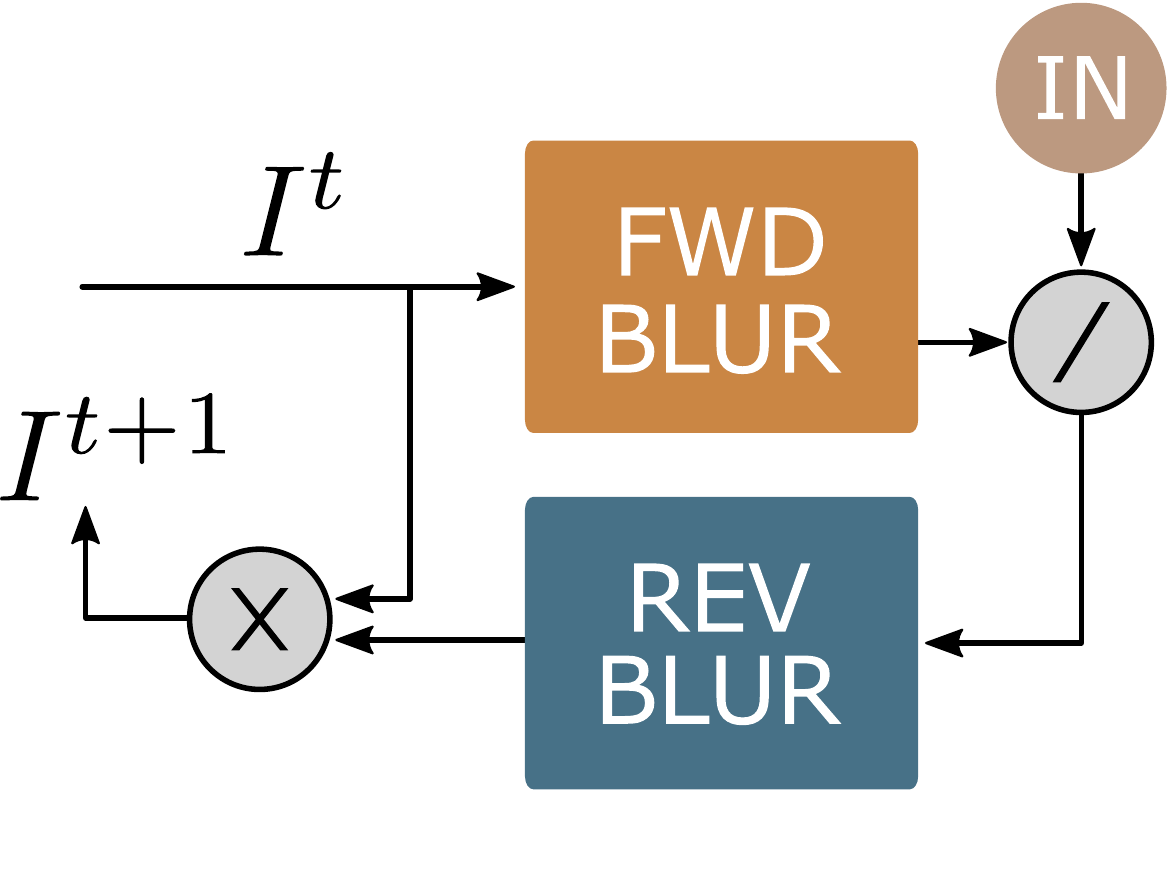}
	\caption{Generalizing the Richardson-Lucy algorithm by replacing convolution with light field rendering of motion blur.}
	\label{Fig_RLSchematic}
\end{figure}

In this work we introduce a method for deblurring light fields of arbitrary 3-D geometry and under arbitrary camera motion.  The proposed approach is a generalization of the \gls{RL} deblurring algorithm~\cite{richardson1972bayesian,lucy1974iterative} in which 2-D convolution is replaced with light field rendering.  The resulting algorithm, depicted in Fig.~\ref{Fig_RLSchematic}, employs light field interpolation to render novel views and simulate motion blur --~no model of the scene's geometry is employed. Our approach is elegant and non-obvious, as all previous attempts at LF motion deblurring have arrived at very different and severely limited solutions compared with ours. Ours is the first example, to our knowledge, of a method dealing with nonplanar scenes and 6-\gls{DOF} camera motion without explicitly estimating scene geometry.

We show results for rendered light fields and light fields captured using a commercially available lenslet-based camera.  Quantitative experimental results require repeatable and known camera trajectories, for which we employ an industrial robot arm capable of sweeping the camera through arbitrary 6-\gls{DOF} trajectories.  Extensive qualitative and quantitative results confirm the method operates robustly over a range of geometry and camera motion, including commonly occurring cases for which previously published methods fail. 

We include a detailed and insightful mathematical proof that the algorithm converges to the maximum-likelihood estimate of the unblurred scene under Poisson noise.  We also introduce a novel regularization term enforcing equal parallax motion in vertical and horizontal dimensions which, combined with previously published regularization based on total variation, significantly improves deblurring results.

Complex 3-D scenes generally yield blur that varies in direction and magnitude on a per-pixel basis, complicating the use of 2-D methods and requiring expensive per-pixel motion models.  The proposed method requires only a description of the camera's trajectory, which for short exposure times is well approximated by a 6-D constant-velocity vector. Its low dimensionality makes the proposed method less computationally complex in the case of known camera motion, and attractive for generalization to blind deconvolution.

As with conventional Richardson-Lucy deconvolution, the proposed method is not blind.  However, we expect this work to form the basis for blind deblurring, e.g.~by following the generalization of 2-D Richardson-Lucy to blind deconvolution~\cite{fish1995blind}.  

The proposed method has some important limitations: it assumes motion blur caused by camera motion or by rigid motion of scene elements, and without extension will not deal with blur induced by relative motion between scene elements.  We also assume knowledge of the camera's motion as acquired from an \gls{IMU} or visual odometry, and require that the light field camera be calibrated and its imagery rectified to allow accurate rendering.  Blind deblurring, in which the motion of the camera and the deblurred image are jointly estimated, is left as future work.  Although the tradeoff between exposure time and light gathering has been addressed in the context of exposure manipulation~\cite{raskar2006coded, agrawal2009invertible}, we focus here on the possibilities offered by light field cameras with conventional exposure regimes.

\section{Related Work}

Classic deblurring approaches operate in 2-D, assuming a constant blur kernel across the image~\cite{richardson1972bayesian,lucy1974iterative}. In general, however, nonuniform apparent motion due to complex scene geometry results in highly variable motion blur.  Adapting to such scenarios requires varying the blur kernel across the image, a process equivalent to estimating the scene's geometry.

Moving beyond two dimensions, Tai et al.~\cite{tai2011richardson} demonstrate a modification of the \gls{RL} deblurring algorithm~\cite{richardson1972bayesian, lucy1974iterative} to incorporate planar projective motion. Their method outperforms spatially invariant blur kernels, though it deals poorly with scenes exhibiting large depth variations, as these break the planar motion assumption. We extend this work using light field rendering and regularization techniques, lifting the planar motion assumption and correctly handling arbitrary scene geometry. 

Joshi et al.\ address spatially varying blur by instrumenting the camera with an \gls{IMU}~\cite{joshi2010image}.  This improves deconvolution by providing an initial camera motion estimate, but their method imposes a constant-depth assumption making it inappropriate for scenes with large depth variations.

Xu and Jia~\cite{xu2012depth} address depth variation by performing depth estimation from a stereo camera. The depth estimate is broken into layers, and these drive a set of \gls{PSF} estimates. Their method requires two cameras and explicit depth estimation from blurry input images, and limits processing to a set of layers, rather than dealing naturally with smooth depth variation.

Levin~\cite{levin2006blind} presents a blind method that segments images based on the statistics of image derivatives, and deblurs each segment with a 1D blur kernel estimate. Because it is based on segments, the method does not deal well with the continuously varying blur commonly associated with smooth depth variation or camera motion.  The rich relationship between camera motion, 3-D scene structure and blur shape is ignored.

Chandramouli et al.~\cite{chandramouli2014light} address blind deconvolution of light fields with decimated spatial sampling. Their method approximates the scene as 2-D, assumes a Lambertian scene, and is not easily extended to handle depth variation. Our method by contrast operates correctly on 3-D scenes with spatially varying blur, and is not limited to Lambertian scenes.

Snoswell and Singh decompose the blurred light field into discrete planes in a process akin to the discrete focal stack transform~\cite{snoswell2014light, perez2014fast}. Each depth plane is independently deblurred, then recombined based on a global depth estimate. This technique relies on forming an accurate depth estimate from the blurred input image, and this fails in low-texture areas and for large amounts of blur. The per-plane deblurring is carried out using 2-D deconvolution, limiting the method to in-plane camera motion. Our method, by contrast, operates directly on the input light field, does not rely on a depth estimate, and works with 6-\gls{DOF} camera motion.

We employ regularization based on anisotropic total variation, which has previously appeared in various forms~\cite{heber2013variational, goldluecke2013variational, wanner2013variational}.  We also propose equiparallax regularization, enforcing equal rates of apparent motion in pairs of light field dimensions. To our knowledge, this form of regularization has not been previously published.

Concurrently with our work, Srinivasan et al.~\cite{srinivasan2017light} introduced a blind deblurring algorithm that jointly estimates the deblurred light field and the camera's trajectory. Although it does not handle camera rotation, it may be possible to generalize their method to handle 6-\gls{DOF} motion.

\section{Plenoptic Richardson-Lucy}


The \gls{RL} deblurring algorithm~\cite{richardson1972bayesian,lucy1974iterative} is typically expressed in terms of convolutions, as
\begin{equation}
I^{t+1} = I^t \cdot \left( \frac{B}{I^t \otimes k} \otimes \tilde{k} \right),
\end{equation}
where the division and multiplication are element-wise. 
Here $B$ is the blurry input image, $\otimes$ denotes convolution, $k$ is the \gls{PSF} of the blurring process, and $\tilde{k}$ reverses $k$ along each of its dimensions.

In this work, as in~\cite{tai2011richardson}, we generalize the blurring process by replacing the denominator $I^t \otimes k$ with a generic \emph{forward blur} operation, and the second convolution $\otimes \tilde{k}$ with a generic \emph{reverse blur} operation, as depicted in Fig.~\ref{Fig_RLSchematic}. 

We restrict our attention to the case of motion blur induced by camera motion in a static scene, or equivalently by rigid scene motion.  Even under this assumption, blur is conventionally difficult to simulate due to the nonuniform apparent motion associated with 3-D scene geometry. 
To address this we make use of light field rendering to simulate motion blur without estimating or making assumptions about the scene's geometry.

The light field was first introduced to allow efficient rendering of novel views~\cite{levoy1996light}. Because camera motion-induced blur can be simulated as the summation of views along a camera trajectory, light field rendering is easily extended to simulating motion blur. The camera's trajectory $P$ is broken into $N$ individual views, and each view is rendered through any of a range of light field rendering techniques. We employ one of the simplest, quadrilinear interpolation~\cite{levoy1996light}, as it requires no depth estimation. Reverse blurring is achieved by reversing each dimension of the simulated camera trajectory to yield the inverse trajectory $\tilde{P}$. Pseudocode for the resulting algorithm is shown in Fig.~\ref{fig_Pseudocode}.

\renewcommand*\Call[2]{\textproc{#1}(#2)}  
\begin{figure}
\begin{algorithmic}[1]
\Function{Deblur}{$I_0, Path$}
\State $I \gets I_0$
\Loop
\State $I_B \gets \Call{Blur}{I,Path}$
\State $R \gets I_0 / I_B$
\State $R \gets \Call{Blur}{R,\Call{Reverse}{Path}}$
\State $I \gets R I$ \Comment{$I$ converges to deblurred image}
\EndLoop
\EndFunction
\Function{Blur}{$I,Path$}
\State $F \gets 0$
\For{$N$ $Views$ in $Path$}
	\State $F \gets F +$\Call{Render}{$I,View$}
\EndFor
\State \textbf{return} $F$
\EndFunction
\end{algorithmic}
\caption{The Light Field Richardson-Lucy Algorithm}
\label{fig_Pseudocode}
\end{figure}

It is known that when the \gls{RL} algorithm converges it yields the maximum-likelihood estimate of the unblurred scene under Poisson noise~\cite{shepp1982maximum}. In the following section we show that our modified algorithm retains this property.

\subsection{Derivation}
\label{sect_ThePudding}

Here we consider the problem of restoring light fields corrupted by motion blur and 
Poisson noise. 
We study light fields embedded in $\R^4$, whose real-valued intensities are defined 
on a closed domain $\Omega \subseteq \R^4$. 
Let $L: \Omega \mapsto \R^+$ denote the unknown and blur-free light field and let 
$\B: \Omega \mapsto \R^+$ denote the observed light field degraded by motion blur according 
to the following model 
\begin{align}
B(w)= \int_{\mathrlap{w+\gamma_w}} \ L(w) ds, \ \  w \in \Omega.
\end{align}
That is, the measured intensity along a given light ray is the result of integrating the 
light field intensities along the entire trajectory taken by each such ray.
Here $\gamma_w : [0, 1] \mapsto \Omega$ parametrises the path of an individual light ray 
at $w\in \R^4$. 
We only consider regular curves arc-length parametrized of class $C^\infty$, for which 
we have
\begin{align}
\int_{\mathrlap{\gamma_w}} \ ds =1.
\label{arclength}
\end{align}
To avoid boundary effects we let $\Omega = \R^4$. 

Under Poisson noise the conditional probability density function for an individual 
light ray at $w\in \Omega$ is given by 
\begin{align}
P(\B(w)|L,\gamma_w) = \frac{B(w)^{\B(w)} }{\B(w)!}e^{-B(w)}.
\label{likelihood}
\end{align}
For an entire light field the log likelihood can be written
\begin{align}
\L(\B|L,\gamma):= \log \left(
\prod_{w\in \Omega} P(\B(w)|L,\gamma_w)
\right)
 \\
=\int_{\Omega} \B(w) log [B(w)] - B(w) - log [\B(w)!] dw.
\label{loglikelihood}
\end{align}
Note that since $B(w)$ is linear in $L$ it follows 
that $\L(\B|L,\gamma)$ is a concave function. 
Finding $L$ is then stated as the maximum a posteriori estimator of 
\eqref{likelihood}, or equivalently as the maximizer of \eqref{loglikelihood}. 

We can write, 
\begin{align}
L(w)&= \argmax_{L(w)\geq 0} 
\int_{\Omega} 
\underbrace{
\B(w) log [B(w)] - B(w)  
}_{f(L)}
dw.
\label{map} 
\end{align} 

The Lagrange function of \eqref{map} becomes, 
\begin{align}
F(L,\Lambda)= f(L) + \int_\Omega \Lambda(w) L(w) dw, \\
\Lambda: \Omega \mapsto \R^+
\end{align}
and the corresponding KKT-conditions
\begin{align}
\frac{\partial F}{\partial L}(w)  + \Lambda(w) = & 0, \label{kkt0a} \\
L(w) \geq & 0, \\
\Lambda(w) \geq & 0, \\
\Lambda(w) \L(w) = & 0, \ \forall w \in \Omega.
\label{kkt0d}
\end{align}
Or equivalently 
\begin{align}
L(w)\frac{\partial f}{\partial L}(w) = & 0, \ \ \textnormal{if } L(w)> 0
\label{kk2a} \\
\frac{\partial f}{\partial L}(w) \geq & 0, \ \ \textnormal{if } L(w)=0. 
\label{kk2b} 
\end{align}
The partial derivative of $f$ with respect to $L$ becomes
\begin{align}
&\frac{\partial f}{\partial L} = 
\frac{\partial}{\partial L} \left(
\int_{\Omega} \B(w) log [B(w)] - B(w) dw
\right) \\
&=  \int_{\Omega} \frac{\B(w)}{B(w)} 
\frac{\partial}{\partial L} \left( 
\int\displaylimits_{\mathrlap{w+\gamma_w}} \ L(w) ds
\right) - \frac{\partial}{\partial L}
\left( 
\int\displaylimits_{\mathrlap{w+\gamma_w}} \ L(w) ds
\right)
 dw \\
&=  \int_{\Omega} \frac{\B(w)}{B(w)} 
\frac{\partial}{\partial L} \left( 
\int\displaylimits_{\mathrlap{w+\gamma_w}} \ L(w) ds
\right)dw 
 - \int\displaylimits_{\mathrlap{w+\gamma_w}} \ ds \\
 &=  \int\displaylimits_{\mathrlap{w+\gamma^{-}_w}} \frac{\B(w)}{B(w)} dw - 1.
\label{dfdl}
\end{align}
With $\gamma^{-}_w : [0, 1] \mapsto \Omega$ denoting the direction reversal of the curve $\gamma_w$, 
i.e.\ $\gamma^{-}_w(t)=\gamma_w(1-t)$.
The last equality follows from \eqref{arclength}, the arc-length 
parameterization of $\gamma_w$.
Inserting \eqref{dfdl} in \eqref{kk2a} yields 
\begin{align}
L(w)\int\displaylimits_{{w+\gamma^{-}_w}} \frac{\B(w)}{B(w)} ds = L(w).
\label{RL1}
\end{align}
The \gls{RL} algorithm can then be derived as the fixed-point  iteration 
of \eqref{RL1}. 
We arrive at the familiar multiplicative \gls{RL} iteration%
\begin{align}
L^{n+1}(w)= L^n(w)\int\displaylimits_{{w+\gamma^{-}_w}} 
\frac{\B(w)}{\int\displaylimits_{{w+\gamma_w}} \ L^n(w) ds} ds.
\label{RL2_first}
\end{align} 
The convergence of the iteration \eqref{RL2_first} can be established
from the work of \cite{shepp1982maximum}. 
Adhering to the analysis therein, it is straightforward to show 
that $L^{n+1}(w)\geq L^{n}(w)$. 
From the concavity and boundedness of $L(w)$ it 
can then be proven that 
\eqref{RL2_first} will converge to a solution to \eqref{map}.
We refer the reader to \cite{shepp1982maximum} for details. 

%
%
%

\subsection{Regularization}

The inclusion of priors on the light field $L(w)$, in the form of a regularizing term 
$R(w,L(w),\nabla L(w))$, 
into the generalized \gls{RL} iteration \eqref{RL2_first} is as straightforward 
as in preceding work \cite{tai2011richardson}. 
Let \eqref{map} now instead be
\begin{align}
L(w)&= \argmin_{L(w)\geq 0} 
\int_{\Omega} 
\B(w) log [B(w)] - B(w) ] \nonumber \\
&+ \rho R(L(w))
dw,
\label{map_reg} 
\end{align} 
with the constant $\rho$ defining the weight of the regularization term.
The equivalent KKT-condition to \eqref{kk2a} then becomes 
\begin{align}
L(w) 
\Bigg[
\frac{\partial f}{\partial L}(w) 
+ \rho 
\underbrace{
\left( 
\frac{\partial R}{\partial L}(w)-
\nabla \cdot \frac{\partial R}{\partial \nabla L}(w)
\right)
}_{E(w)}
\Bigg]
= & 0.
\label{kk3a} 
\end{align}
Using \eqref{dfdl} we can write \eqref{kk3a} as 
\begin{align}
L(w)\int\displaylimits_{{w+\gamma^{-}_w}} \frac{\B(w)}{B(w)} ds = (1-\rho E(w))L(w),
\end{align}
arriving at the regularized variant of the multiplicative \gls{RL} 
iteration for light fields,
\begin{align}
L^{n+1}(w)= \frac{L^n(w)}{1-\rho E(w)}
\int\displaylimits_{{w+\gamma^{-}_w}} \frac{\B(w)}{\int\displaylimits_{{w+\gamma_w}} \ L^n(w) ds} ds.
\label{RL2_second}
\end{align} 

\subsubsection{Anisotropic Total variation}

Regularization by total variation is well established as a means of suppressing image noise amplification by minimizing the magnitude of gradients in the deblurred image~\cite{chan1998total, dey2004deconvolution}.  We employ a generalization to 4-D total variation for light fields, including anisotropy introduced to reflect the limited range of epipolar slopes typical of light fields~\cite{heber2013variational, goldluecke2013variational, wanner2013variational}.

Here we consider smoothed anisotropic total variation regularizers of the 
following form,
\begin{align}
R_{tv}(\nabla L)=\int_\Omega 
\sqrt{ \nabla L(\omega)^T D \nabla L(\omega) + \epsilon} \ d\omega,
\end{align}
with $\epsilon > 0$ and where 
directional sensitivity is described by the 
positive definite tensor $D \in \mathcal{S}^{4\times 4}_{++}$.  
For this choice of regularizing term, $E(\omega)$ in \eqref{kk3a}, 
then becomes 
\begin{align}
E_{tv}(\omega)= \nabla \cdot \frac{D \nabla L(\omega)}
{\sqrt{ \nabla L(\omega)^T D \nabla L(\omega) + \epsilon}}.
\end{align}

\subsubsection{Equiparallax}

It is well established that epipolar slopes in horizontal and vertical light field dimensions must be equal --~this is a consequence of apparent motion occurring at the same rate across horizontal and vertical camera positions.  The consequences of this ``equiparallax'' have been exploited to formulate highly selective noise rejecting filters for light fields in the frequency domain~\cite{dansereau2015linear}.  In this work, we construct a regularization term that enforces the equiparallax constraint in order to further suppress noise amplification and to enforce valid light field geometry in the deblured imagery.

In \cite{dansereau2015linear} it was shown that for Lambertian scenes without occlusion boundaries the following constraints on the partial derivatives of the light field 
must hold, 
\begin{align}
\frac{\nabla_s L (w)}{\nabla_u L (w)}=
\frac{\nabla_t L(w) }{\nabla_v L(w)},
\end{align}
with $\nabla_u L,\nabla_v L \neq 0$, and the dimensions $s,t,u,v$ following the well-known two-plane light field parameterization~\cite{levoy1996light}.  From this we derive the regularizer
\begin{align}
R_{ep}(\nabla L)&=  \int_\Omega \sqrt{ g(\omega)^2 + \epsilon} \ d\omega, \\
g(\omega) &= \nabla_s L(\omega) \nabla_v L(\omega) - \nabla_u L(\omega) \nabla_t L(\omega),
\end{align}
resulting in an $E(\omega)$ as in \eqref{kk3a} given by
\begin{align}
E_{ep}(\omega)= \nabla \cdot
\frac{g(\omega)}{\sqrt{g(\omega)^2+\epsilon} }
\matris{\nabla_v L(\omega) \\ -\nabla_u L(\omega) \\ - \nabla_t L(\omega) \\ \nabla_s L(\omega)}.
\end{align} 

\begin{figure*}
	\centering
	\subfloat[IN $T_x$ (32.1~dB)]{\includegraphics[width=0.5\columnwidth]{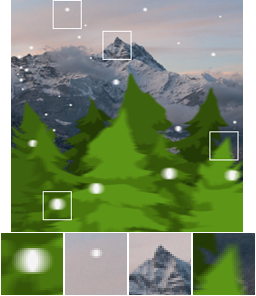}}\hfil
	\subfloat[LF-RL $T_x$ (33.6~dB)]{\includegraphics[width=0.5\columnwidth]{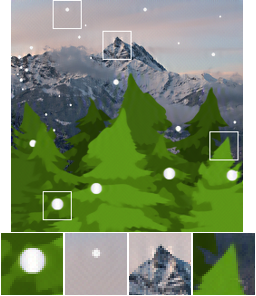}}\hfil
	\subfloat[2D-RL $T_x$ 9~pix (22~dB)]{\includegraphics[width=0.5\columnwidth]{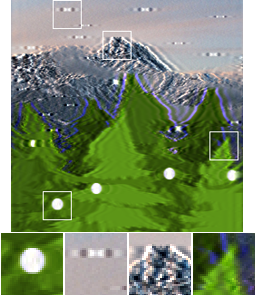}}\hfil
	\subfloat[2D-RL $T_x$ 5~pix (25.4~dB)]{\includegraphics[width=0.5\columnwidth]{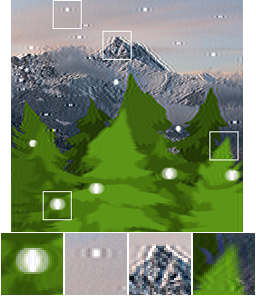}}\\\vspace{-0.5em}
	\subfloat[2D-RL $T_x$ 2~pix (30.7~dB)]{\includegraphics[width=0.5\columnwidth]{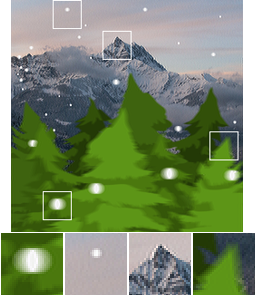}}\hfil
	\subfloat[Wiener $T_x$ 9~pix (20.2~dB)]{\includegraphics[width=0.5\columnwidth]{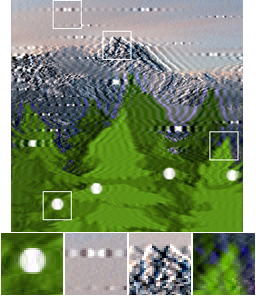}}\hfil
	\subfloat[Wiener $T_x$ 5~pix (23.8~dB)]{\includegraphics[width=0.5\columnwidth]{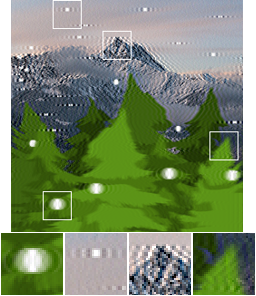}}\hfil
	\subfloat[Wiener $T_x$ 2~pix (30.2~dB)]{\includegraphics[width=0.5\columnwidth]{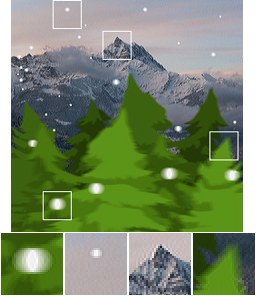}}\vspace{0.5em}
	\caption{(a) Translation in $x$ yields scene-dependent spatially varying motion blur. (b) The proposed algorithm converges on a deblurred result without forming an explicit scene model. (c-h) 2-D methods cannot handle the spatially varying blur, with 9-, 5- and 2-pixel kernels each only addressing subsets of the image.}
	\label{Fig_SnowSceneResults_x}
\end{figure*}

\begin{figure}
	\centering
	\subfloat[IN $R_y$ (26.9~dB)]{\includegraphics[width=0.5\columnwidth]{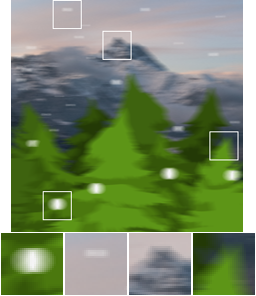}}\hfil
	\subfloat[LF-RL $R_y$ (32.1~dB)]{\includegraphics[width=0.5\columnwidth]{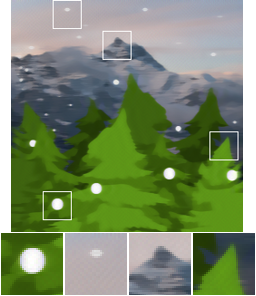}}\\\vspace{-0.5em}
	\subfloat[2D-RL $R_y$ (29.2~dB)]{\includegraphics[width=0.5\columnwidth]{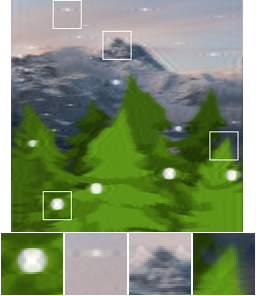}}\hfil
	\subfloat[Wiener $R_y$ (28.6~dB)]{\includegraphics[width=0.5\columnwidth]{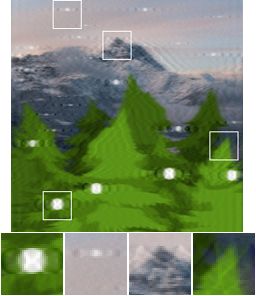}}\vspace{0.5em}
	\caption{(a) Rotation about $y$ yields spatially invariant blur, and is therefore well addressed by (b) our method and (c,d) 2-D deconvolution methods.  We suspect the strong regularization afforded by the light field explains our method's superior results.}
	\label{Fig_SnowSceneResults_wy}
\end{figure}

\begin{figure}
	\centering
	\subfloat[IN $R_z$ (29.2~dB)]{\includegraphics[width=0.5\columnwidth]{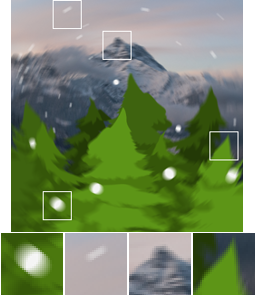}}\hfil
	\subfloat[IN $T_z$ (32.6~dB)]{\includegraphics[width=0.5\columnwidth]{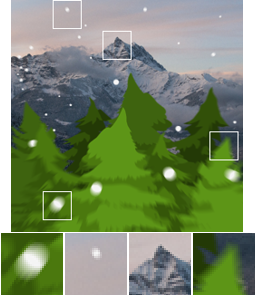}}\\\vspace{-0.5em}
	\subfloat[LF-RL $R_z$ (33.8~dB)]{\includegraphics[width=0.5\columnwidth]{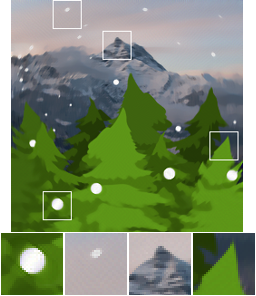}}\hfil
	\subfloat[LF-RL $T_z$ (32.6~dB)]{\includegraphics[width=0.5\columnwidth]{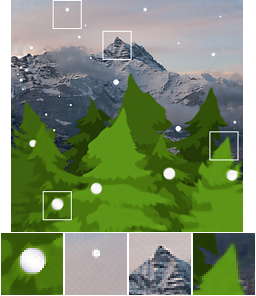}}\\\vspace{-0.5em}
	\subfloat[PROJ-RL $R_z$ (23.4~dB)]{\includegraphics[width=0.5\columnwidth]{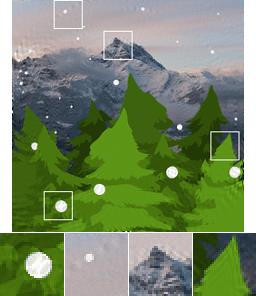}}\hfil
	\subfloat[PROJ-RL $T_z$ (16.9~dB)]{\includegraphics[width=0.5\columnwidth]{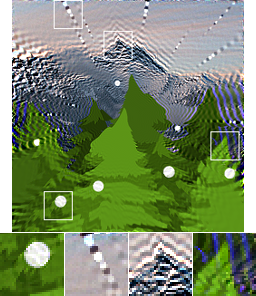}}\\\vspace{0.5em}
	\caption{(a) Rotation about $z$ yields spatially varying but scene-independent blur, while (b) translation along $z$ yields spatially varying scene-dependent blur; (c-d) our method is capable of dealing with both these cases, while 2-D methods cannot; (e) Projective \gls{RL} deals correctly with rotation about $z$, but not (f) translation along $z$, due to the scene-dependent blur.}
	\label{Fig_SnowSceneResults_z}
\end{figure}

\section{Experiments}

\subsection{Implementation Details}

Because we are working with relatively short exposure durations, the camera's trajectory can be well approximated using a constant velocity given by the vector $v = [T_x,T_y,T_z, R_x,R_y,R_z]$.  Although this limits our implementation of the proposed method to constant-velocity cases, the method is more general in that it is capable of handling any camera trajectory $P(n)$ that can be approximated as a set of $N$ discrete poses.

The constant-velocity assumption allows for a simplification in the blur processes:  For a trajectory defined over the unit time step, we set the pose $P(-0.5) = -v/2$ and $P(0.5) = v/2$.  This fixes the deblurred image to the center of the trajectory, i.e.~$P(0)$, and allows the use of identical forward and reverse blur operations. 

An important parameter of the deblurring process is the number of steps $N$ to take in approximating the camera's trajectory.  Unless otherwise stated, all experiments employed $N=10$ steps.  Also important are the number of iterations over which of the RL algorithm is run.  We found that most deblurring occurred within the first ten iterations, but that there was occasionally improvement up to 50 iterations, especially with regularization enabled.  In general, results are shown for 50 RL iterations.  Regularization was employed with an equiparallax gain of $\rho_{ep}=0.05$, and a total variation gain of $\rho_{tv}=0.01$ with an anisotropy of 8 favouring edges in the $u,v$ dimensions.

\subsection{Rendered Scenes}

We begin by establishing the ability of the algorithm to deal with complex 3-D geometry under different types of camera motion.  For this we employed a raytracer\footnote{\url{http://dgd.vision/Tools/LFSynth}} to generate a variety of scenes and simulated camera motions.  Motion blur was simulated during the raytracing process by integrating views along a camera trajectory. This was done during light field creation using conventional raytracing techniques, ensuring that motion blur simulation was not carried out using the light field rendering process built into the deblurring algorithm.

The rendered light fields have $15 \times 15 \times 256 \times 256 \times 3$ samples, for which our unoptimized MATLAB implementation took about 2 minutes per iteration on an 8-core i7-4790 CPU at 3.60~GHz.

We identified four characteristic motion classes, depicted in Figs.~\ref{Fig_SnowSceneResults_x}--\ref{Fig_SnowSceneResults_z}, and for each class we compared the output of the proposed algorithm with relevant competing methods. Note that all displayed results correspond to the central view of the light field.  To facilitate discussion we assign $z$ as the optical axis of the camera, with $x$ pointing to the right and $y$ up.  Numerical results are the error relative to an unblurred view of the scene, taken as $-20 \log_{10}(\mathit{RMSE})$, for a maximum pixel magnitude of 1.

The first motion class is translation in the $x, y$ plane, for which parallax motion yields a variety of effective blur magnitudes, and prevents the effective application of conventional deblurring algorithms.  Shown in Fig.~\ref{Fig_SnowSceneResults_x} is translation in $x$, correctly deblurred by the proposed method, but only partially deblurred by 2-D methods which must be tuned to specific subsets of the image. Shown are examples for 9-, 5- and 2-pixel 2-D blur kernels, corresponding roughly to the blur lengths of the inset scene features.  Note that projective \gls{RL}~\cite{tai2011richardson} would also fail here because the scene is not well approximated as a plane.

The second motion class is nodal rotation excluding rotation about $z$, yielding approximately constant projected motion throughout the scene.  Note that nodal rotation is not possible throughout the entire light field, due to the spatial extent of the camera array, but we are visualizing the central view of the light field for which nodal rotation is possible. Shown in Fig.~\ref{Fig_SnowSceneResults_wy} is rotation about the vertical axis, $R_y$, yielding constant blur throughout the central view.  As seen in the figure, both the proposed method and conventional 2-D deblurring algorithms correctly deblur this scene.

The third and fourth motion classes are rotation about and translation along $z$. The former yields geometry-independent blur, which can be well addressed by the projective deblurring algorithm of Tai et al.~\cite{tai2011richardson}, or by a scene-independent spatially varying 2-D deconvolution.  The latter, translation about $z$, yields scene-dependent blur similar to translation in $x,y$.  Examples of these two motion classes are depicted in Fig.~\ref{Fig_SnowSceneResults_z}, with the proposed algorithm correctly handling both.  No meaningful results can be obtained from spatially invariant 2-D deconvolution in these cases, but projective \gls{RL} deals correctly with the rotational case, as seen in the bottom row.  Note that the  projective \gls{RL} implementation we used yielded a rotational offset which we removed in order to maximize the numerical performance.

\begin{figure*}
	\centering
	\subfloat[IN $T_x$ (21.7~dB)]{\includegraphics[width=0.5\columnwidth]{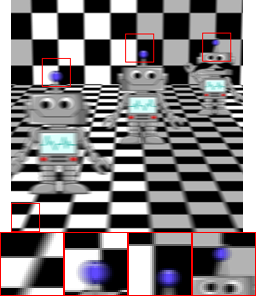}}\hfil
	\subfloat[IN $R_y$ (19.8~dB)]{\includegraphics[width=0.5\columnwidth]{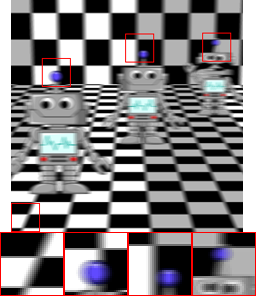}}\hfil
	\subfloat[IN $T_z$ (19.4~dB)]{\includegraphics[width=0.5\columnwidth]{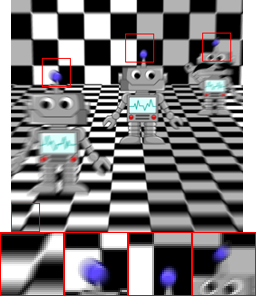}}\hfil
	\subfloat[IN $R_z$ (16.9~dB)]{\includegraphics[width=0.5\columnwidth]{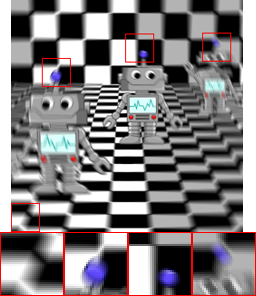}}\\\vspace{-0.5em}
	\subfloat[LF-RL $T_x$ (27.5~dB)]{\includegraphics[width=0.5\columnwidth]{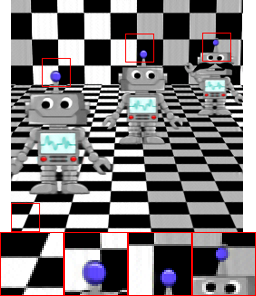}}\hfil
	\subfloat[LF-RL $R_y$ (26.1~dB)]{\includegraphics[width=0.5\columnwidth]{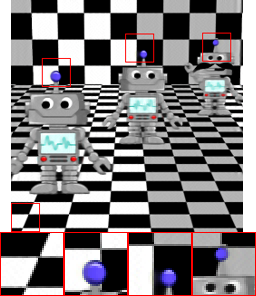}}\hfil
	\subfloat[LF-RL $T_z$ (23.3~dB)]{\includegraphics[width=0.5\columnwidth]{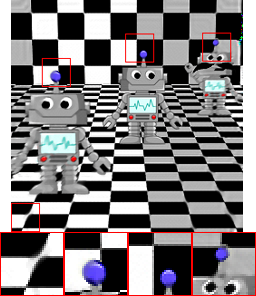}\label{Subfig_CompareRegularization}}\hfil
	\subfloat[LF-RL $R_z$ (22.4~dB)]{\includegraphics[width=0.5\columnwidth]{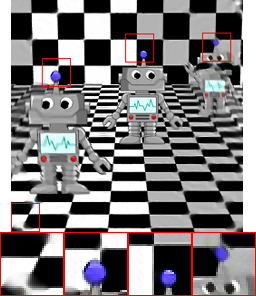}}\\\vspace{0.5em}
	\caption{Validating the proposed method over four classes of motion: (a) translation in $x,y$, (b) nodal rotation in $x,y$, (c) translation in $z$ and (d) rotation about $z$. (e-f) The proposed method has correctly dealt with all cases including spatially varying and scene-dependent blur, even in the presence of occlusions.}
	\label{Fig_RoboSceneResults}
\end{figure*}

\begin{figure}
	\centering
	\includegraphics[width=0.5\columnwidth]{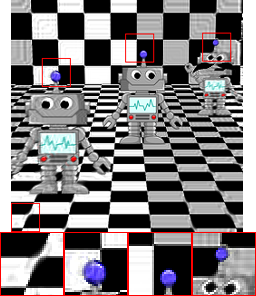}
	\caption{For the same light field depicted in Fig.~\ref{Subfig_CompareRegularization}, omitting regularization yields characteristic ringing and noise amplification (22.0~dB).}
	\label{Fig_Regularization}
\end{figure}

Results for the proposed method are shown for a second rendered scene in Fig.~\ref{Fig_RoboSceneResults}.  Noteworthy is that for all characteristic motion classes, and across complex scene geometry, the proposed method was able to correctly deblur the scene.  We expect this to hold for arbitrary combinations of motion classes, subject to the limits of motion discussed in following sections.  The impact of omitting regularization is shown in Fig.~\ref{Fig_Regularization}.

\subsection{Robot-Mounted Camera Experiments}

\begin{figure}
	\centering
	\includegraphics[width=0.9\columnwidth]{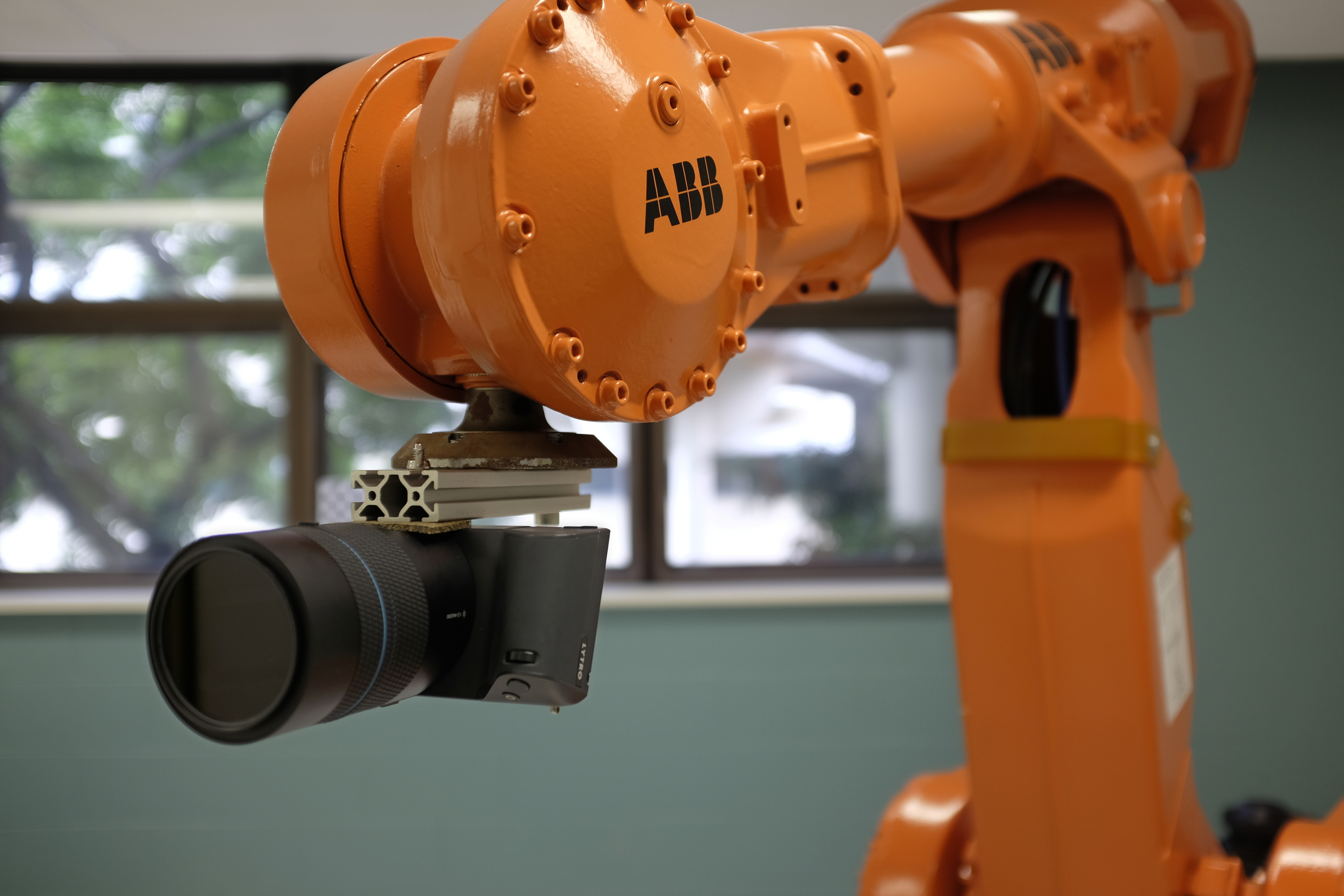} 
	\caption{A commercially available light field camera mounted on an industrial robot arm produces repeatable and ground-truthed 6-\gls{DOF} camera motion.}
	\label{Fig_CameraOnArm}
\end{figure}

The proposed method requires knowledge of the camera's motion --~development of a blind method is expected to be possible following the generalization of 2-D Richardson-Lucy to blind deconvolution, and is left as future work.  As such, to validate the method on real-world imagery, we mounted a commercially available lenslet-based light field camera~-- a Lytro Illum~-- on an industrial robot arm, as depicted in Fig.~\ref{Fig_CameraOnArm}.  The arm was programmed for a range of motion classes and rates, including the four characteristic ones described in the previous section.

\begin{figure*}
	\centering
	\subfloat[IN $T_x$]{\includegraphics[width=0.66\columnwidth]{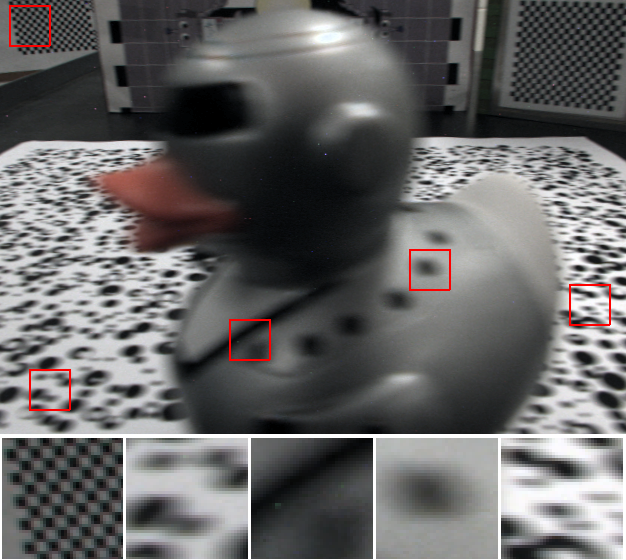}}\hfil
	\subfloat[IN $T_z$]{\includegraphics[width=0.66\columnwidth]{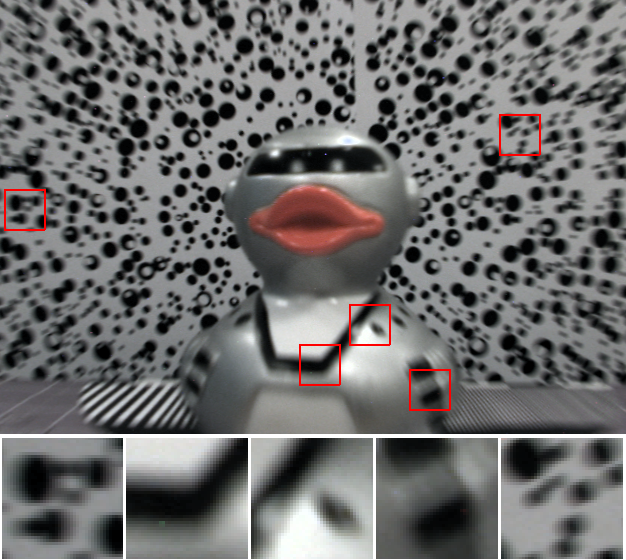}}\hfil
	\subfloat[IN $R_z$]{\includegraphics[width=0.66\columnwidth]{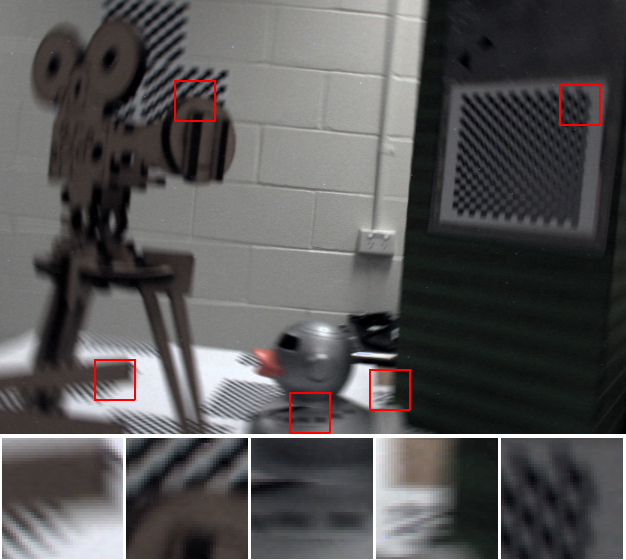}}\\\vspace{-0.5em}
	\subfloat[LF-RL $T_x$]{\includegraphics[width=0.66\columnwidth]{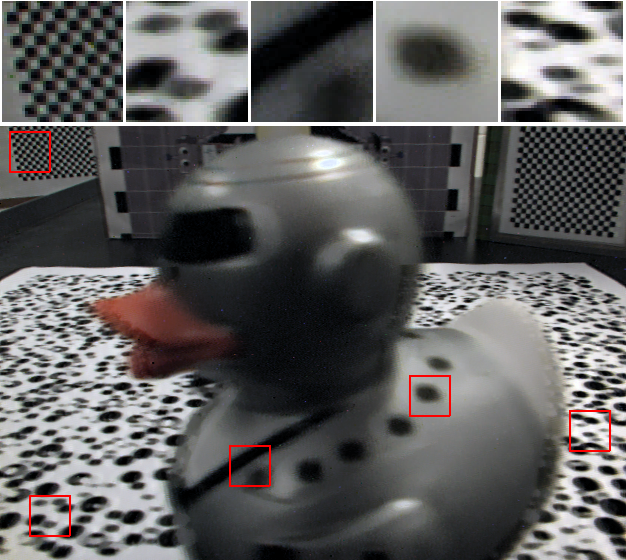}\label{Subfig_OcclusionIssues}}\hfil
	\subfloat[LF-RL $T_z$]{\includegraphics[width=0.66\columnwidth]{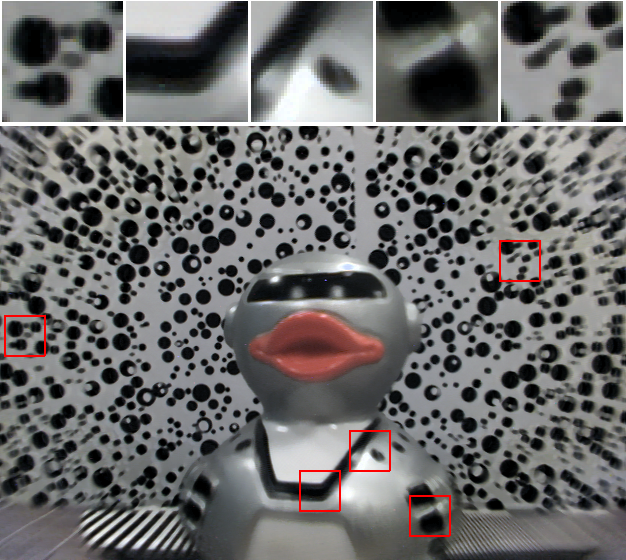}}\hfil
	\subfloat[LF-RL $R_z$]{\includegraphics[width=0.66\columnwidth]{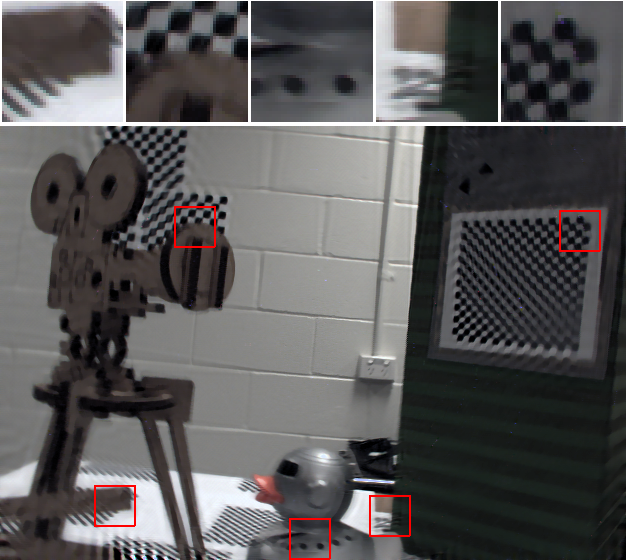}}\\\vspace{0.5em}
	\caption{Imagery captured with a commercially available light field camera mounted on an industrial robot arm, showing (a) translation in $x$, (b) translation along $z$ and (c) rotation about $z$. (d-f) In all cases the method has reduced visible blur, dealing correctly with scene-dependent and spatially varying blur.  We attribute the lower performance near the edges of (e) to edge effects associated with this relatively large camera motion.  Note the marked improvement in the robo-ducky's textural details, and the checkerboard details in (f).}
	\label{Fig_ABBResults}
\end{figure*}

\subsubsection{Calibration}

Accurate rendering of motion blur requires calibrated and rectified light fields.  This was accomplished using the Light Field Toolbox for Matlab~\cite{dansereau2013decoding}.  We found it necessary to exclude a border of 2 pixels near lenslet edges during the calibration process, due to limitations of the lens distortion model employed in the toolbox.  We fixed the camera's zoom to its widest field of view, and selected the hyperfocal distance as the focal setting.

The rectified light fields have $15 \times 15 \times 626 \times 434 \times 3$ samples, though we discard a border of 1 pixel yielding a total of 13 samples rather than 15 in the first two dimensions.  Because these are much larger than the rendered light fields, runtime was longer, with our unoptimized MATLAB implementation taking about 5 minutes per iteration on an 8-core i7-4790 CPU at 3.60~GHz.

The arm was programmed by setting two endpoints for each motion class, and the arm was set to linearly oscillate between them over a range of velocities. For translation in $x,y$ and rotation about $x, y$, we found the imagery to be relatively insensitive to slight errors in the arm's movement. For rotation and translation about $z$, however, we found small errors in the arm's motion yielded several-pixel deviations from the ideal.  For these types of motion, each path endpoint was manually adjusted to maintain a fiducial at the center of the image, resulting in close-to-ideal imagery.

Example light fields measured using the arm-mounted camera, and the corresponding deblurred light fields, are shown in Fig.~\ref{Fig_ABBResults}. The leftmost example shows horizontal motion at 75~mm/sec over a 1/10~sec exposure; the center example shows translation towards the scene at 75~mm/sec over a 1/5~sec exposure; and the final example shows rotation about $z$ at 0.6545~rad/sec over a 1/10~sec exposure.  Note the recovery of edge detail, especially in the robo-ducky's texture, and the checkerboards in the rightmost image. Note also the artifacts near occlusions in (d), for which further investigation is indicated.

\begin{figure}
	\centering
	\subfloat[Blurry]{\includegraphics[width=0.45\columnwidth]{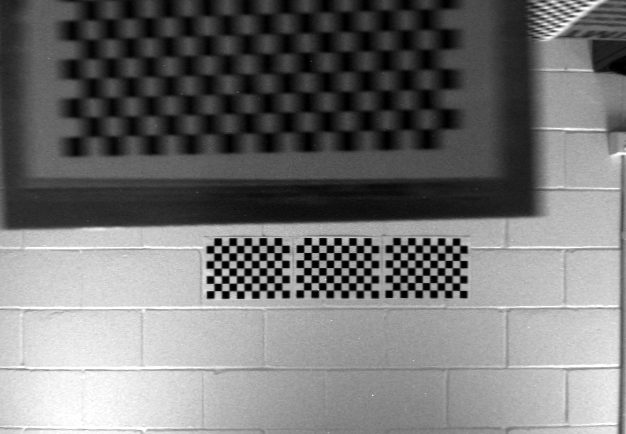}\label{Subfig_Blurcheck_Blurry}}\hfil
	\subfloat[Still]{\includegraphics[width=0.45\columnwidth]{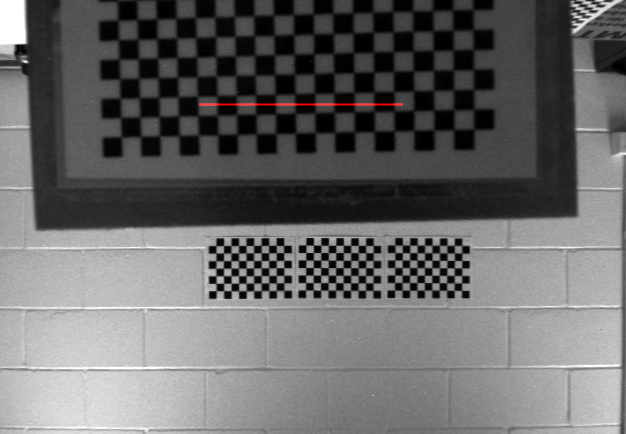}\label{Subfig_Blurcheck_Still}}\\\vspace{-0.5em}
	\subfloat[Simulated Blur]{\includegraphics[width=0.45\columnwidth]{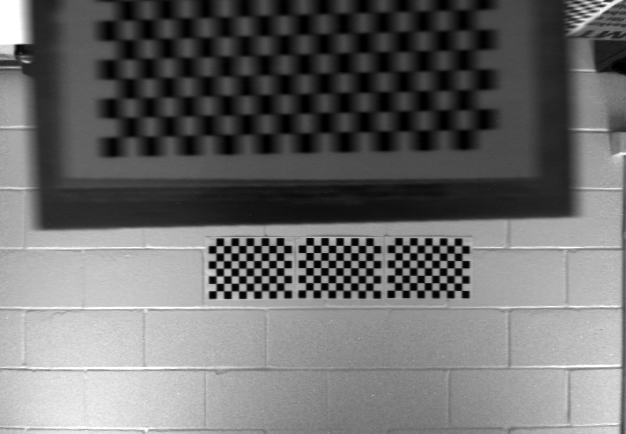}\label{Subfig_Blurcheck_Render}}\hfil  
	\subfloat[Deblurred LF-RL]{\includegraphics[width=0.45\columnwidth]{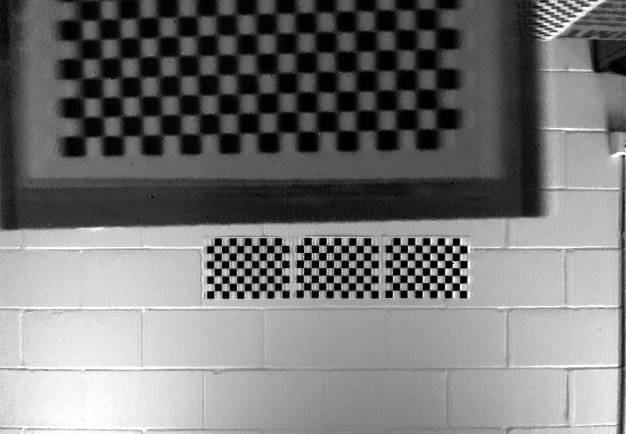}\label{Subfig_Blurcheck_LFRL}}\\\vspace{-0.5em}
	\subfloat[Deblurred 2D-RL 20~pix]{\includegraphics[width=0.45\columnwidth]{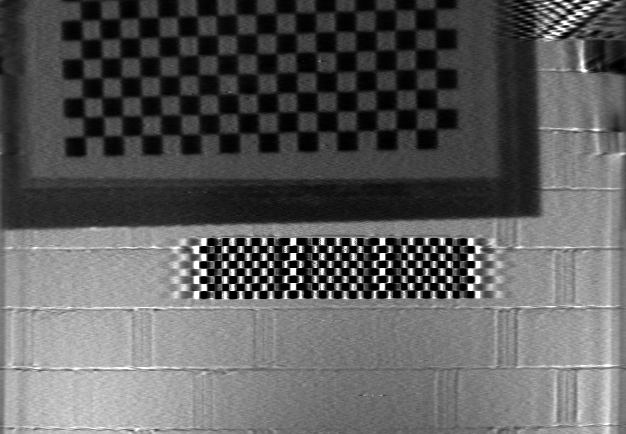}}\hfil
	\subfloat[Deblurred 2D-RL 2~pix]{\includegraphics[width=0.45\columnwidth]{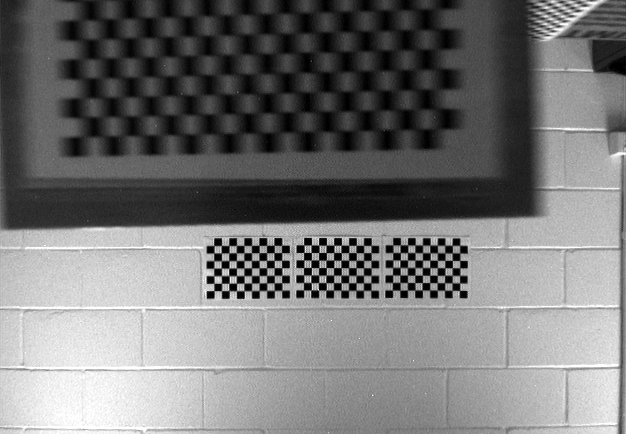}}\vspace{0.5em}
	\caption{Validating calibration of the camera and arm: (a) a series of blurry images is paired with (b) still views of the same; (c) simulating blur from the still image using the model-free light field blur simulation confirms correct calibration of the camera and arm, and operation of the blur process.  See Fig.~\ref{Fig_CheckSimBlurTraces} for intensity plots of (a-c), and Fig.~\ref{Fig_EdgeVsVelResults} for plots of edge energy over a range of camera velocities. (d-f) Depict deblurring results using the proposed method and 2D-RL tuned to 20- and 2-pixel blur kernels, respectively; These results confirm the efficacy of the proposed method compared with 2-D methods, which do not address scene-dependent blur.}
	\label{Fig_BlurCheckExamples}
\end{figure}

\subsubsection{Validating Calibration and Rendering}

As a means of validating the arm and camera calibration, we collected images of scenes over a range of camera velocities, paired with still frames of the same.  The motion blur simulation step was then applied to the still frames, and compared with the corresponding measured blur. This doubles as validation of the light field rendering of motion blur on which the proposed method relies.

The arm was set to travel between 0 and 100~mm/sec, for an exposure time of 1/20\textsuperscript{th} of a second.  The scene included two checkerboards, one a few cm from the camera, and one about 3~m away.  Examples of a blurry image, corresponding to an arm velocity of 75~mm/sec, and stationary view, are shown in Figs.~\ref{Subfig_Blurcheck_Blurry} and~\ref{Subfig_Blurcheck_Still}.  The still frame was passed to the motion blur simulation, producing the result shown in Fig.~\ref{Subfig_Blurcheck_Render}.  Visually, this is a close match to the behaviour seen in the directly observed blur: the foreground checkerboard shows similar blur levels, while the background checkerboard remains mostly unchanged. 

\begin{figure}
	\centering
	\includegraphics[width=0.7\columnwidth]{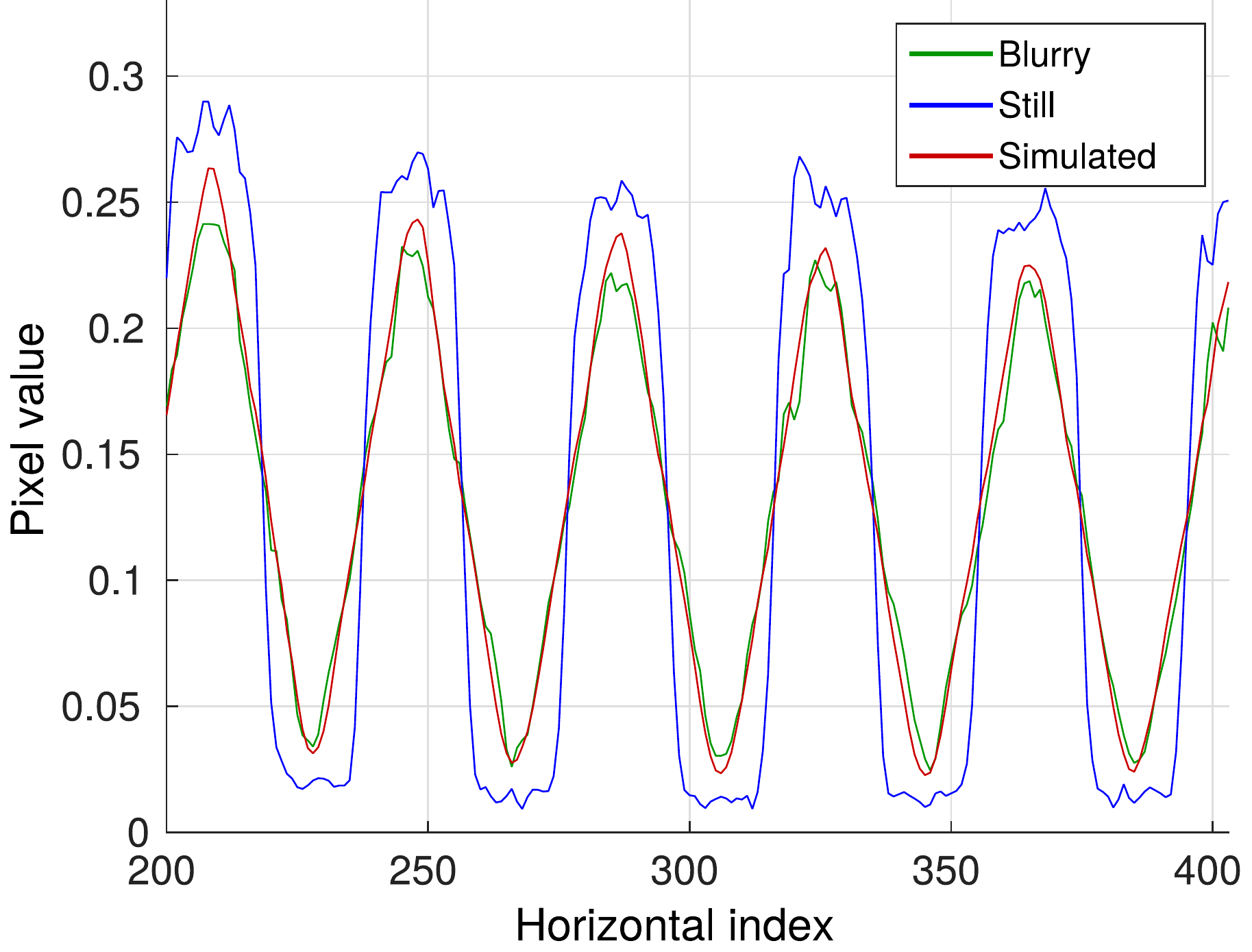}
	\caption{Intensity plot along the still, blurry and simulated blur images shown in Fig.~\ref{Fig_BlurCheckExamples}. The location of the plot is highlighted in red in Fig.~\ref{Subfig_Blurcheck_Still}.  There is good agreement between the simulated blur and measured blur, confirming correct calibration of the camera and arm velocity.  Quantification over a range of velocities is shown in Fig.~\ref{Fig_EdgeVsVelResults}.}
	\label{Fig_CheckSimBlurTraces}
\end{figure}

As confirmation of the simulated blur extent, intensity plots of the measured blur, still frame, and simulated blur from Fig.~\ref{Fig_BlurCheckExamples} are shown in Fig.~\ref{Fig_CheckSimBlurTraces} --~the extent of the plot is depicted in red in Fig.~\ref{Subfig_Blurcheck_Still}.  Note that the still frame shows relatively sharp edges, while the measured and simulated blur show virtually identical shapes.  The measured blur trace has been shifted horizontally to align with the simulated blur, because the still and blurry images were not measured from exactly the same locations.

This step also allowed us to experimentally establish the range of motion possible in deblurring.  The camera's effective baseline and field of view limit the available range of simulated translation and rotation. Starting from a still frame collected in the previous step, we simulated blur over increasing values of translation and rotation, observing the extents of the light field as seen in the central image.   We found maximum translations in $x,y$ of up to 3.95~mm and in $z$ of up to 7.5~mm, beyond which the field of view narrowed significantly.  For rotations, we found that up to 0.06~rad in $x$ or $y$ resulted in a loss of less than 20~pixels at the image border, with larger rotations causing larger borders.  Rotation about $z$ is effectively unlimited, though image edges tend to be impacted due to the non-square aspect ratio.


\begin{figure}
	\centering
	\subfloat[]{\includegraphics[width=0.8\columnwidth]{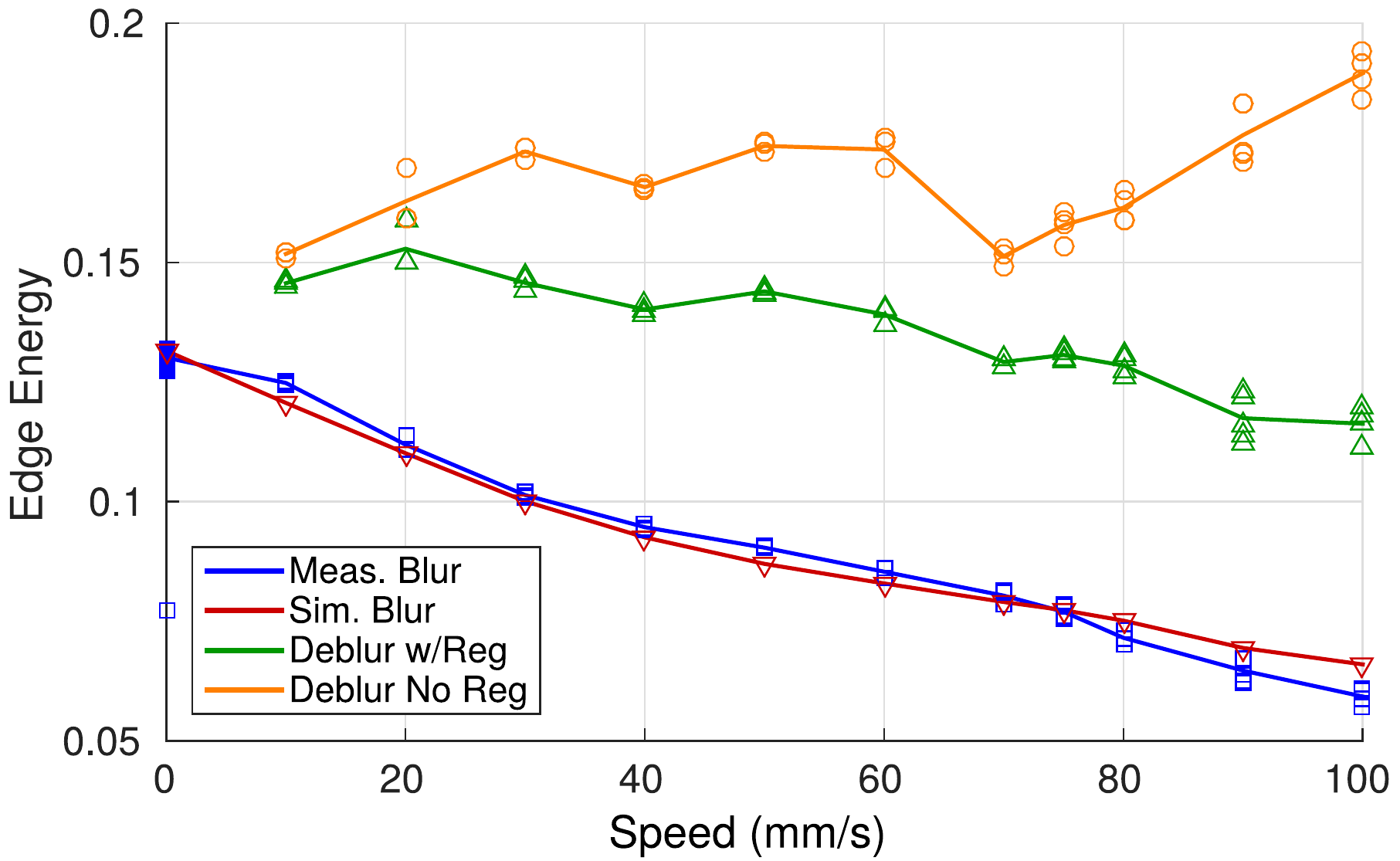}}\\\vspace{-0.5em}
	\subfloat[]{\includegraphics[width=0.8\columnwidth]{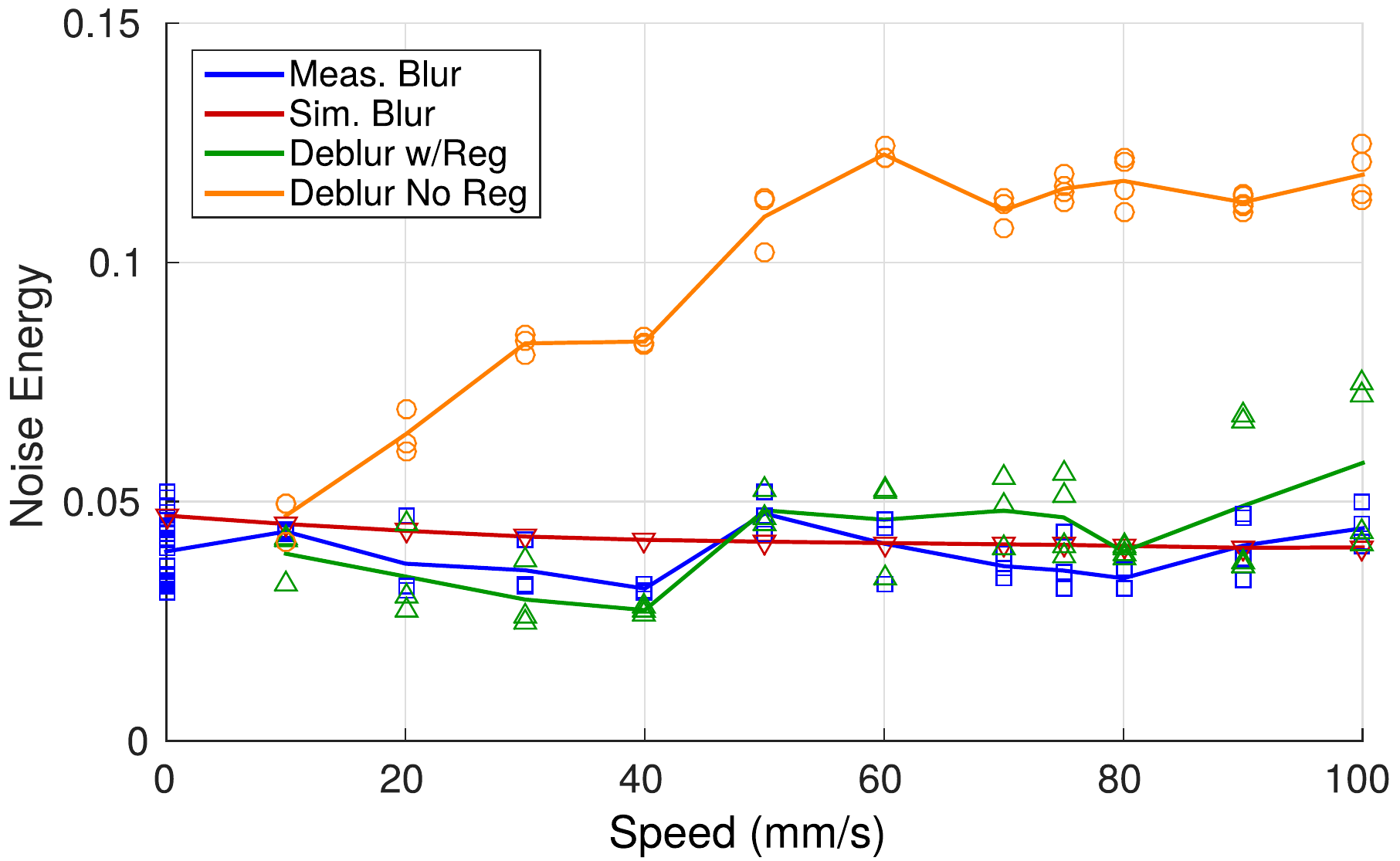}}\vspace{0.5em}
	\caption{Quantifying performance of the proposed method: the checkerboard experiment depicted in Fig.~\ref{Fig_BlurCheckExamples} was repeated over a range of camera velocities, measuring (a) edge energy in the checkerboard, and (b) standard deviation in the white regions adjacent to the checkerboard.  Blue and red traces show good agreement between measured and simulated motion blur.  Green traces establish consistent improvement of the imagery using the proposed method without dramatic amplification of noise or introduction of ringing, while the unregularized results in orange show a dramatic increase in noise and ringing.  Note that the blur simulation is limited to 79~mm/sec, explaining the decrease in performance above that speed.}
	\label{Fig_EdgeVsVelResults}
\end{figure}

We repeated the blur simulation experiment over a range of velocities, measuring edge energy and noise / ringing content. Edge energy was taken over the visible area of the closer checkerboard pattern, as the mean of the square of the first difference in the horizontal direction.  Noise energy was measured as the standard deviation over a $16 \times 160$ pixel white patch adjacent to the checkerboard.  The results are shown as red and blue traces in Fig.~\ref{Fig_EdgeVsVelResults}, with between 3 and 5 image repetitions of each nonzero velocity image. Note that the edge energy in measured and simulated-blur images match closely, while the noise level is about constant for both.  Because of the motion limits discussed above, velocities beyond 79~mm/sec are not well represented in the blur simulation, and this is reflected in the deviation in edge energy seen above that speed.

\subsubsection{Deblurring Performance}

We applied the proposed algorithm, both with and without regularization, to the checkerboard images gathered in the blur validation experiment.  An example of the output is shown in Fig.~\ref{Subfig_Blurcheck_LFRL}.  Though not perfect, it is clear that all elements of the scene have been treated correctly, with a significant reduction in visible blur. A 2-D \gls{RL} algorithm was also tested, for blurs of 20 and 2 pixels, and as seen in the bottom row of Fig.~\ref{Fig_BlurCheckExamples} this resulted in favourable results for foreground or background elements, but not both.

Numerical results for the proposed method are shown in the green and orange traces in Fig.~\ref{Fig_EdgeVsVelResults}, again with between 3 and 5 repetitions per image.  The green trace corresponds to light field \gls{RL} with total variation and equiparallax regularization, while the orange omits the regularization stage.  Although the non-regularized method has yielded more edge energy, it has also increased the noise level --~this is essentially the amplification of noise and the introduction of edge artifacts characteristic of unregularized deblurring.  The regularized result, on the other hand, does not appreciably increase the noise level for velocities below 80~mm/sec, but does significantly improve edge content.  This lines up well with a qualitative assessment of the results.

\section{Conclusions and Future Work}

We presented a method for deblurring light fields of arbitrary 3-D scenes with arbitrary camera motion.  This is the first published example, to our knowledge, of an algorithm capable of dealing with 3-D geometry and 6-\gls{DOF} camera motion without requiring an explicit 3-D model of the scene.

We introduced a novel regularization term enforcing equal rates of apparent motion in horizontal and vertical light field dimensions, and included a mathematical proof that the algorithm converges to the maximum-likelihood estimate of the unblurred scene under Poisson noise.

A commercially available lenslet-based camera mounted on a robot arm gave us precise control of the camera's motion, allowing validation of the method on real-world imagery.  Both qualitative and quantitative results over rendered and real-world imagery confirmed the efficacy of the method over a range of camera motions.

The method relies on prior, accurate knowledge of the camera's trajectory, and so generalization to blind deconvolution is an obvious next step.  The extremely low dimensionality of the blur model --~limited to six numbers in the case of a constant-velocity trajectory --~makes promising the possibility of an optimization-based blind deconvolution algorithm.

Validation in the presence of speculars and transparency would be interesting.  Because light field rendering deals correctly with these elements, we expect the method to perform well in their presence.  A detailed analysis of the regularization parameters would also be useful.

Some interesting limitations arose in validating the method, most noteworthy being undesirable patterns arising near occlusion boundaries, e.g. in Fig.~\ref{Subfig_OcclusionIssues}.  It is unclear whether this is the result of miscalibration of the camera's velocity or optics, rendering artifacts due to the use of quadrilinear interpolation, or whether this reflects a fundamental limitation of the method.

Finally, a promising line of work could combine the methods explored here with other ideas from computational imaging, in particular modulated exposure regimes like flutter shutter~\cite{raskar2006coded}.

\textbf{Acknowledgments} 
This research was supported by the Australian Research Council through the Centre of Excellence for Robotic Vision (project number CE140100016) and grant DE130101775. Computational resources and services provided by the HPC and Research Support Group, Queensland University of Technology, Brisbane, Australia.

{\small
\bibliographystyle{ieee}
\bibliography{LightFieldBib}
}

\end{document}